\pdfoutput=1
\documentclass[11pt]{article}

\usepackage[]{acl}

\usepackage{times}
\usepackage{latexsym}

\usepackage[T1]{fontenc}
\usepackage[utf8]{inputenc}

\usepackage{microtype}

\usepackage{enumitem}
\usepackage{graphicx}
\usepackage[export]{adjustbox}
\usepackage{booktabs}
\usepackage{float}
\usepackage{subcaption}
\usepackage{tabularx}
\usepackage{colortbl}

\newcommand{\FitB}{\textsc{FitB}}

\newcommand{\FITB}{\textsc{FillBlank}}
\newcommand{\FITE}{\textsc{Cont}}
\newcommand{\FITBFITE}{\textsc{FillBlankCont}}

\newcommand{\cFITB}{\textsc{C4FillBlank}}
\newcommand{\cFITE}{\textsc{C4FillEnd}}

\newcommand{\rwpFITB}{\textsc{RwpFillBlank}}
\newcommand{\rwpFITE}{\textsc{RwpFillEnd}}

\newcommand{\rocFITB}{\textsc{RocFillMiddle}}
\newcommand{\rocFITE}{\textsc{RocFillEnd}}

\newcommand{\LLM}{\textsc{Llm}}

\title{The Case for a Single Model that can Both\\Generate Continuations and Fill in the Blank}
\author{
Daphne Ippolito\textsuperscript{1,2}\ \ \ \ 
Liam Dugan\textsuperscript{1}\ \ \
Emily Reif\textsuperscript{2}\\
\textbf{Ann Yuan\textsuperscript{2}}\ \ \ \ 
\textbf{Andy Coenen\textsuperscript{2}}\ \ \ \  
\textbf{Chris Callison-Burch\textsuperscript{1}\ \ }\\
\textsuperscript{1}University of Pennsylvania \ \ \
\textsuperscript{2}Google Research\\
 \{{\tt daphnei, ldugan, ccb\}@seas.upenn.edu}
 \\
 \{{\tt ereif, annyuan, andycoenen\}@google.com}\\
}
\date{}

\begin{document}
\maketitle
\begin{abstract}
The task of inserting text into a specified position in a passage, known as fill in the blank (\FitB), is useful for a variety of applications where writers interact with a natural language generation (NLG) system to craft text.
While previous work has tackled this problem with models trained specifically to do the fill-in-the-blank task, a more useful model is one that can effectively perform \textit{both} \FitB{} and continuation.
In this work, we evaluate the feasibility of using a single model to do both tasks.
We show that models pre-trained with a \FitB-style objective are capable of both tasks, while models pre-trained for continuation are not.
Finally, we show how \FitB{} models
can be easily finetuned to allow for fine-grained control over the length and word choice of the generation.
\end{abstract}

\section{Introduction}
% Main contributions
% We are the first to propose using the same model for both fill-in-the-blank and continuation.
% First to compare against strategies which do not require any fine-tuning.
% First to show how models trained on general data transfer to out-of-domain data.

Natural language generation systems are increasingly being incorporated into applications where a human writer and an AI jointly collaborate to construct text.
These range from creative domains such as collaborative story writing \citep{wordcraft,akoury2020storium} to more practical ones such as email composition and code synthesis \citep{buschek2021impact,wu2018smart,austin2021program}.
These applications are often limited to generating text at the end of what has been written so far.
This is because
% both historical language models (LMs) and state-of-the-art neural LMs
language models (LMs)
are typically designed to produce text by repeatedly predicting the next word in a sequence given the previous words.
However, there is a need for more powerful interactive tools which enable writers to solicit insertions at any chosen position within the existing text, a task referred to as fill in the blank (\FitB) or infilling.
For example, a creative writer might want a tool which can insert a description of a place or character, and a programmer might want a system that can fill in the contents of a function located in the middle of their code.

\begin{figure}[t]
    \centering
    \includegraphics[width=0.48\textwidth]{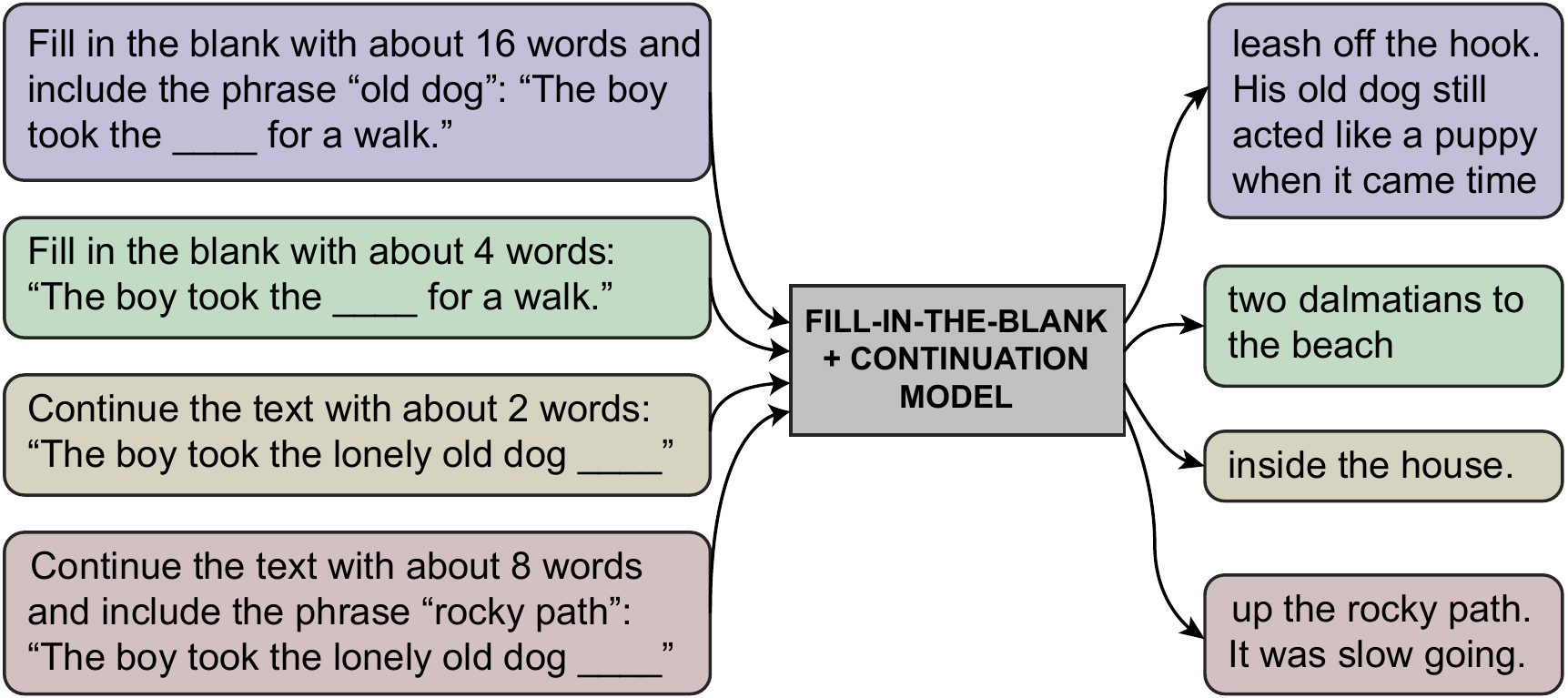}
    \caption{A single model that can handle a variety of related writing tasks is more efficient than separate models per task.}
    \label{fig:leading_figure}
    \vspace{-1em}
\end{figure}

Most prior work tackling \FitB{} consider it a separate task from continuation, one to be specifically optimized for, for example training a model from scratch \citep{ippolito2019unsupervised,zhu2019text} or finetuning a model trained originally for continuation \citep{donahue2020enabling}.
However, having separate trained models for \FitB{} and for continuation is inefficient for downstream applications where maintaining multiple neural networks can be prohibitive.

Any model that can do \FitB{} can be made to do continuation simply by placing the blank at the end of the input.
Thus, in this work we describe how models trained on \FitB{} can be employed effectively for both infilling and continuation operations.
We show how T5 \citep{raffel2019exploring}, one of the most popular pre-trained models, can reasonably handle both tasks, as it was pre-trained with a \FitB-like objective.
Finetuning T5 further improves its ability and also allows for the incorporation of controllability of generation length and word choice.

\begin{table*}[t]
  \centering
  \small
    \begin{tabular}{p{0.12\textwidth}p{0.57\textwidth}p{0.20\textwidth}}
    \hline
    Example Type & Input & Target \\
    \hline
     \cFITB{} no goal & fill: I love avocados. I ate a sandwich covered in them. {\bf \_8\_} I talked to my doctor about it later. It turned out I was allergic to avocados. & After I ate it, my mouth was itchy and tingly. \\
    \hline
     \cFITB{}  with goal & fill: I love avocados. I ate a sandwich covered in them.  {\bf \_8\_} I talked to my doctor about it later. It turned out I was allergic to avocados.  {\bf Goal: mouth was itchy} & After I ate it, my mouth was itchy and tingly. \\
    \hline
     \cFITB{} no goal& fill: I love avocados. I ate a sandwich covered in them. After I ate it, my mouth was itchy and tingly. I talked to my doctor about it later.  {\bf \_8\_}  & It turned out I was allergic to avocados. \\
    \hline
     \cFITE{} with goal & fill: I love avocados. I ate a sandwich covered in them. After I ate it, my mouth was itchy and tingly. I talked to my doctor about it later.  {\bf \_8\_}   {\bf Goal: allergic to} & It turned out I was allergic to avocados. \\
    \hline
    \end{tabular}
   \vspace{-0.6em}
  \caption{Examples of the finetuning objectives. ``8'' is the approximate length in words of the target sequence. During finetuning, about 25\% of training examples took each of these formats.}
  \label{tab:task_examples}
  \vspace{-1em}
\end{table*}%

\section{Supporting \FitB{} and Continuation}
\label{section:methods}
\paragraph{Definitions.}
We define filling in the blank as the task of predicting text to replace a single missing span, usually demarcated with a special token, in an input text passage. (Some prior work considers inputs with multiple blanks, but inserting text at one position at a time better matches the kinds of edits humans do.)
We define continuation in the traditional language modeling sense--predicting the next token in a sequence given only the previous tokens.
\citet{donahue2020enabling} discuss how language modeling is a special case of infilling, and they use this as justification to finetune a continuation-based language model to do infilling.
However, we argue that if continuation is a subtask of infilling, it makes more sense to go in the opposite direction: prioritize a model which can do infilling and check that it achieves satisfactory performance at continuation.

\paragraph{Using a model pre-trained for \FitB.} T5 is a model pre-trained with a ``span corruption'' objective very similar to \FitB; the model is asked to reconstruct the missing text after random sub-sequences of the input are replaced with special identifiers.
Thus, a pre-trained T5 model can be used without any further training to do both continuation and infilling by appropriately choosing text to mask out.
The encoder-decoder architecture of T5 is also more conducive to \FitB{} than the decoder-only architectures that are typically used for continuation-based generation, such as GPT-2 \citep{radford2019language}.
This is because the attention mechanism in encoder-decoder architectures allows the context on the left side of the blank to attend to the context on the right, while decoder-only architectures only support masked attention (each token can only attend to the positions to its left).

Even though T5's pre-training objective was a form of \FitB, finetuning is still advantageous.
For one, our definition of \FitB{} involves only a single masked out substring, not multiple, so finetuning improves alignment with the goal task.
% Furthermore,
Finetuning also allows us to incorporate additional conditioning signals not supported by the pre-trained T5, such as being able to specify the desired length of the generated text or specify words that ought to be included in the blank, a task we refer to as ``goal conditioning.''
Length control, which comes by default in a traditional language model
%since one can simply sample
by simply sampling more or fewer
tokens, is particularly necessary for \FitB, where the end of the generation must fit seamlessly with the text to its right.

\paragraph{Using a model pre-trained for continuation.}
The biggest language models available today were largely trained in the continuation rather than the \FitB{} paradigm \citep{gpt3,gpt-neo}.
Since our primary goal is to have a single model for both tasks, we also address the question: if a continuation-trained model is big enough, can it handle \FitB{} without the need for finetuning?
Few-shot learning with large language models, as popularized by \citet{gpt3}, has had success on many tasks in NLP.
We try out this approach for \FitB{} by designing a few-shot prompt containing several demonstrations of the \FitB{} task, formulated in a similar ``infilling by language modelling" template to that proposed by \citet{donahue2020enabling}.
Further details on our approach to selecting a few-shot prompt are in Appendix \ref{appendix:fs_prompt_selection}.

\section{Experiments}
\label{section:experiments}
\paragraph{Model.}
For all experiments with T5, we use the 800M parameter v1.1 `large' model (
Appendix \ref{appendix:further_ft_exp} gives additional results from the 3B parameter `XL' model).
To finetune T5 for \FitB, we construct training examples from documents by first partitioning the document text into a left context, gap, and right context.
The input to the model is then the left and right contexts concatenated with textual representations of the additional conditioning signals.
The target sequence is the true text for the blank.
This formulation easily supports continuation, as the blank can be deliberately placed at the end (i.e., providing no right context).
Finetuning examples are drawn from C4, the same dataset T5 was pre-trained on.
Documents are split into word sequences, and these are then randomly truncated to be between 256 and 512 words long.
A substring of between 1 and 64 words is selected to be blanked out.
For half of the finetuning examples, the location of the blank is randomly selected, and for the other half, it is always placed at the end.
To support length conditioning, we follow \citet{roberts2020exploring} and include a bucketed version of the target length as part of the blank.
To support goal conditioning, for half the examples, a random substring of up to half the words of the target is appended to the end of the input.
Examples are shown in Table \ref{tab:task_examples}.

\paragraph{Baselines}
We compare T5 against \citet{thoppilan2022lamda}'s 137B parameter decoder-only language model (referred to in this paper as \LLM{}).
Trained explicitly for continuation, this model has been used successfully for few-shot learning in other domains \citep{austin2021program,reif2021recipe}.
%This model is used (1) as a standard continuation model, passing in only the left context of an example as the prompt; and (2) in a few-shot learning paradigm.
We use the \LLM{} in two ways: (1) as a standard continuation model, prompting with only the left context of an example; and (2) in a few-shot learning paradigm.

\paragraph{Evaluation Datasets}
We evaluate continuation and \FitB{} on C4 as well as two story writing datasets. We chose this domain because creative writing assistant applications are one of the key areas we expect to benefit from multi-task models \citep{wordcraft}.
Reddit Writing Prompts (\textsc{Rwp}) is a corpus of stories from the `r/WritingPrompts' sub-Reddit \citep{fan2018hierarchical}, and we construct validation sets \rwpFITB{} and \rwpFITE{} using the same method described in the previous section.
We cap the C4 and \textsc{Rwp} validation sets to 5,000 examples each.
% We also include a third validation set \rwpFITS, where gaps are randomly chosen but always exactly one sentence long.
ROC Stories (\textsc{Roc}) is a crowd-sourced dataset of five-sentence commonsense stories \citep{mostafazadeh2016corpus}.
For ROC Stories, the 2018 validation set is used to construct \rocFITB, where the middle sentence of each story is blanked out, and \rocFITE, where the last sentence is blanked out.
Unless otherwise noted, all evaluation is done without goal conditioning and uses random sampling with top-$k$=50 as the decoding strategy.
Example generations for all evaluation sets can be found at \url{https://bit.ly/2U0Ixxa}.

\begin{table}[t]
\small
\centering
    \begin{tabular}{l|rrr}
    \toprule
    %  & \cFITB & \rocFITB & \rwpFITB & \rwpFITB-Sent \\
    & \textsc{C4Fill} & \textsc{RwpFill} & \textsc{RocFill} \\ % & \textsc{RwpFill} \\
    & \textsc{Blank} & \textsc{Blank} & \textsc{Middle} \\ % & \textsc{Blank}-Sent \\
    % \cline{2-5}
    \hline
    {Few-shot \LLM} & 14.14 & 19.48 & 18.21 \\ % & 16.36 \\
    % {Pre-trained T5 XL} & 10.27 & 13.94 & 21.75 \\ % & TODO \\
    {Pre-trained T5} & 10.38 & 14.08 & 22.62 \\ % & 10.49 \\
    \hline
    % {\FITBFITE{} XL} & 11.49 & 15.05 & 24.87 \\ % & 10.33 \\
    {Finetuned T5} & 10.33 & 14.08 & 20.47 \\ % & 10.37 \\
    {\citet{donahue2020enabling}} & N/A & N/A & 23.28 \\ % & N/A \\
    
    \hline
    {Groundtruth} & 9.41 & 12.99 & 16.90 \\

    \bottomrule
    \end{tabular}
    \vspace{-0.7em}
    \caption{Perplexity of evaluation sets according to \LLM{} when the blank has been filled with approaches involving no fine-tuning (top), finetuned approaches (middle), and the groundtruth (bottom).
    Lower values indicate that the text was considered more fluent by the \LLM{}.
    \label{tab:generative_ppl_results}
    }
\end{table}

\begin{table}[t]
    \centering
    \small
    \begin{tabular}{l|rrr}
    \toprule
    %  & \cFITB & \rocFITB & \rwpFITB & \rwpFITB-Sent \\
    & \textsc{C4Fill} & \textsc{RwpFill} & \textsc{RocFill} \\
    & \textsc{End} & \textsc{End} & \textsc{End}  \\
    % \cline{2-5}
    \hline
    {\LLM} & 9.34  & 12.82 & 15.55 \\
    {Pre-trained T5}  & 10.09 & 13.51 & 21.71 \\
    \hline
    % {\FITBFITE{} XL}  & 10.77 & 14.66 & 22.66 \\
    {T5 \FITBFITE{}} & 10.04 & 13.74 & 19.60 \\
    {T5 \textsc{Lm-Adaption}} & 10.06  & 13.71 & 19.68 \\
    \hline
    {Groundtruth} & 9.41 & 12.99 & 16.90 \\
    \bottomrule
    \end{tabular}
    \vspace{-0.7em}
    \caption{Perplexity of continuation-based evaluation sets when a continuation has been generated using approaches with no finetuning (top) and two settings of finetuning T5 (middle).}
    \label{tab:generative_ppl_continuation_results}
    \vspace{-1em}
\end{table}

\begin{figure}[t]
    \centering
    \includegraphics[width=0.49\textwidth, trim={0 0.55cm 0 0}, clip]{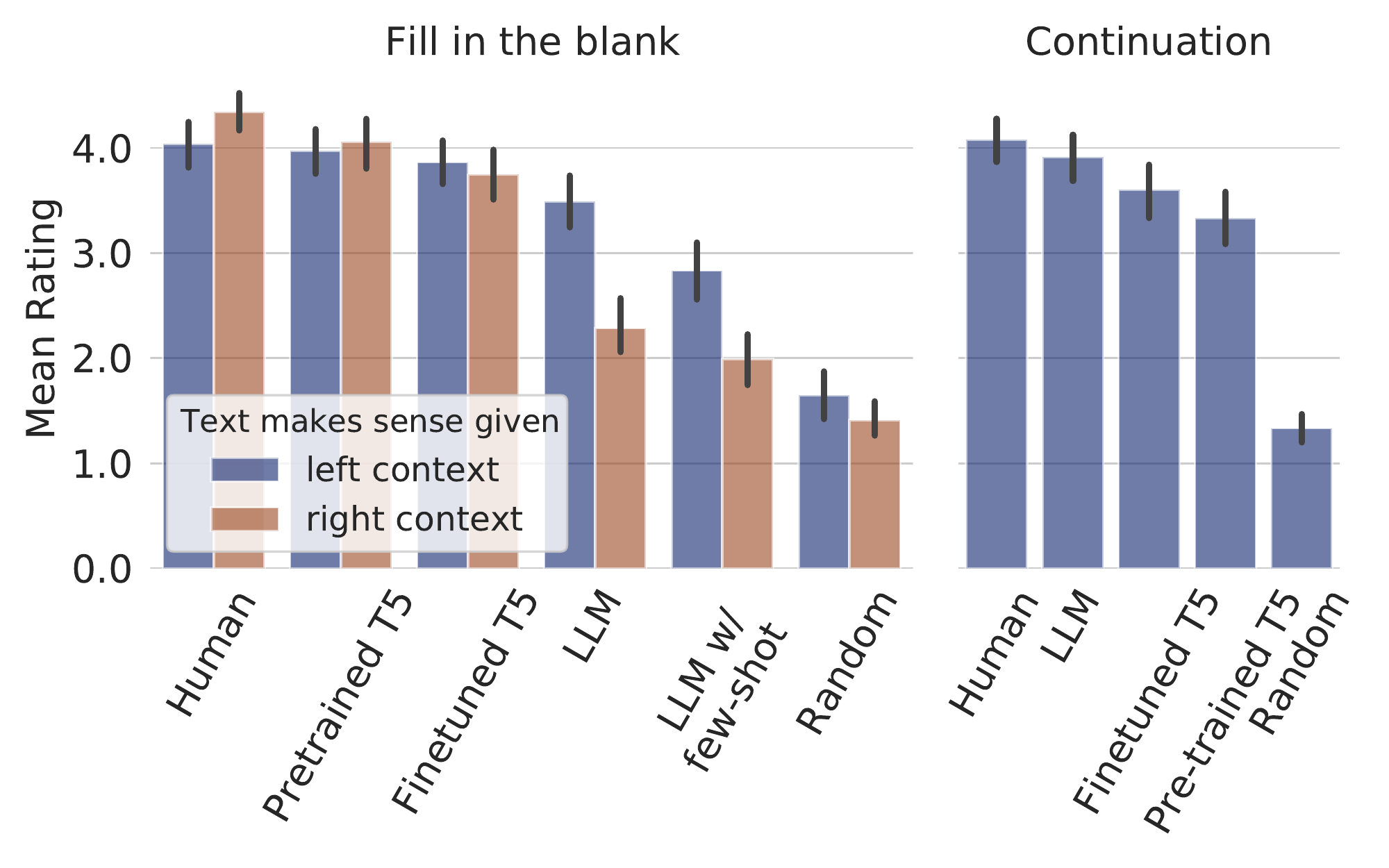}
    \vspace{-1.75em}
    \caption{Human ratings of \FitB{} generations (left) and continuation generations (right). Error bars are 95\% confidence intervals.}
    \label{fig:human_eval_results}
\end{figure}

\begin{table}[t]
\centering
\small
    \begin{tabular}{p{7em}|rr}
    % & \multicolumn{2}{c|}{XL} \\
    \toprule
    % & \multicolumn{2}{c|}{Large} \\
    \textbf{Finetuned T5} & Context & Length \\
     \hline
    \cFITB & 0.860 & 0.877 \\
    \rwpFITB & 0.797 & 0.881 \\
    \hline
    \cFITE & 0.858 & 0.775 \\
    \rwpFITE & 0.791 & 0.746 \\
    \bottomrule
    \end{tabular}
    \vspace{-0.5em}
    \caption{Accuracy of models finetuned on \FITBFITE{} at correctly using provided length and goal conditioning signals. \label{tab:conditioning_signal}}
    \vspace{-1.2em}
\end{table}

\section{Findings}
\paragraph{Automatic Evaluation}
We measure the fluency of proposed generations by evaluating the perplexity of each dataset's examples when the predicted text is placed in the blank \citep{donahue2020enabling}.
We use the \LLM{} to measure perplexity\footnote{Note, since this is the same model being used for generation for our continuation baseline, this metric may be biased.}.
The results are shown in Table \ref{tab:generative_ppl_results}.
We see that the \LLM{} struggles to generate fluent infills, even when used in a few-shot setting.
The only exception to this is ROC Stories, a dataset with fairly simplistic, predictable language.
Finetuning T5 does not result in significantly improved fluency over the pre-trained model except on ROC Stories. 
Lastly, for ROC Stories, we compare against \citet{donahue2020enabling}'s finetuned GPT-2 small, which yielded less fluent predictions.

Table \ref{tab:generative_ppl_continuation_results} shows a similar analysis on our continuation-style datasets.
We see that the pre-trained T5 generates about as fluent continuations as T5 finetuned in the manner described in Section \ref{section:experiments}  (T5 \FITBFITE{}), as well as T5 finetuned for the same number of steps, but only on the continuation task (T5 \textsc{Lm-Adaption}).
The first row of Table \ref{tab:generative_ppl_continuation_results} shows how fluent the \LLM{} scores its own generated continuations.

\paragraph{Human Evaluation}
Human evaluation was conducted on 70 examples, 35 from \rwpFITB{} and 35 from \rwpFITE, with examples about evenly distributed across length buckets.
For \rwpFITB{} evaluation tasks, the rater was presented an input context and several possible sequences that could go in the blank. 
They were asked to rate each sequence first, on how well it fit the text before it, and second, on how well it fit with the text following it, according to a 5-point slider. 
For \rwpFITE{}, the task was almost the same, except that the rater was presented only a left context and asked to rate how well it continued the prompt.
More details are in Appendix \ref{appendix:ft_impl_details}.
Figure \ref{fig:human_eval_results} shows the results.

On the \FitB{} task, the pre-trained and finetuned T5 models were indistinguishable in terms of quality.
The \LLM{} that formed continuations prompted with only the left context did somewhat better than the few-shot \LLM{}, indicating that few-shot learning is not yet a feasible alternative to finetuning.
On the continuation task, the \LLM{} has the highest rating, which is unsurprising since it is a much larger model than T5.
However, the finetuned T5 is rated almost as highly.
Overall, these results suggest that T5, unlike the \LLM{}, can be used effectively for continuation as well as \FitB.
Furthermore, if one doesn't care about controllability, pre-trained T5 can be used effectively for both tasks without any further finetuning.

\definecolor{plotblue}{HTML}{1F78B5}
\definecolor{plotorange}{HTML}{F57E20}
\definecolor{plotgreen}{HTML}{2DA148}

\begin{figure*}[t]
    \centering
    \includegraphics[width=0.99\textwidth]{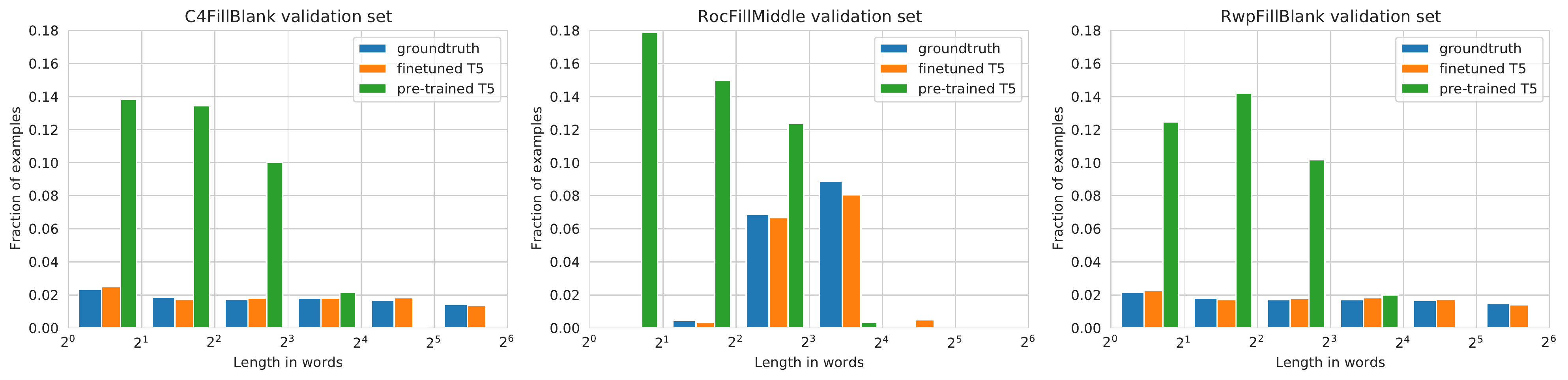}
    \caption{For each of the \FitB{} validation sets, a histogram of the distribution of sequence lengths (measured in words) of the {\textcolor{plotblue}{ground-truth}} blanked out text and the proposed infills from T5 ({\textcolor{plotorange}{after}} and {\textcolor{plotgreen}{before}} finetuning). We see that pre-trained T5 tends to produce text that is shorter than the groundtruth.}
    \label{fig:t5_lengths}
\end{figure*}

\paragraph{Benefits of Controllability}
Despite finetuning not resulting in significantly more fluent outputs, there are still good reasons to finetune; namely, finetuning allows for increased controllability.
For example, length conditioning is extremely important for \FitB models, since it is not possible to control the generation length by simply sampling more or fewer tokens.
Pre-trained T5 tends to produce infill proposals which are shorter than the groundtruth (Figure \ref{fig:t5_lengths}), and there is no way to ask the model to produce longer generations.
In contrast, finetuned T5 was able to produce generations in the target length bucket over 74\% of the time (Table \ref{tab:conditioning_signal}).
% ldugan: This is really interesting, I wonder how many of the prompts were within one bucket of the desired length? I think that number might be a bit more telling
Goal conditioning, while not strictly necessary for either task, has been shown to be useful for generative commonsense reasoning \citep{lin2020commongen} and may empower users in downstream applications such as AI-assisted creative writing \citep{roemmele2021inspiration}. 
Finetuned T5 is able to use all of the specified goal words over 79\% of the time.

\paragraph{Domain Transfer}
Prior work on \FitB{} tends to only evaluate models trained on data from the same domain as the validation set.
Our results show that despite training exclusively on C4, T5 models have strong transferability to more targeted domains such as Reddit Writing Prompts.
This sort of transferability is extremely important for achieving the goal of having a single model which can handle many tasks and domains.

\section{Related Work}
\FitB{} is a form of Cloze task \citep{taylor1953cloze}.
Prior deep-learning approaches to this task include
%\citet{ippolito2019unsupervised} who train a Transformer model from scratch that predicts the blank text given the context, a target length bucket, and a list of tokens which should be included in the prediction, as well as \citet{donahue2020enabling} who finetune GPT-2 \citep{radford2019language} to perform fill-in-the-blank.
training an encoder-decoder model from scratch with length and goal word conditioning
% to predict the blank text given the context, a target length bucket, and a list of tokens which should be included in the prediction
\citep{ippolito2019unsupervised};
finetuning GPT-2 \citep{radford2019language, donahue2020enabling}; and
training a custom self-attention architecture on corrupted text \citep{zhu2019text}.
None of these show that their fill-in-the-blank models remain effective at continuation or perform well on text domains that differ from the training data.
% Our \FitB{} training objective is similar to the ``infilling by language modeling" objective described in \citet{donahue2020enabling}, except since we use an encoder-decoder model instead of a decoder-only model, the attention layers encoding the context in our approach support attending to future tokens positions, not just prior ones.
Related to \FitB, \citet{mori2020finding} investigate a setting where a sentence is randomly deleted from the input, and the model must both predict the location of the deletion as well as its contents.
\citet{huang2020inset} tackle the sentence infilling task using a mixture of BERT and GPT-2.
% to encode the context sentences and GPT-2 to generate the missing sentence given the context's BERT embeddings.
Lastly, many LM pre-training objectives involve masking out parts of the input then predicting the masked values, which is similar to \FitB{} \citep{devlin2018bert,raffel2019exploring,joshi2020spanbert}.
% TODO Cite https://arxiv.org/abs/2008.06048 and other music works once we have more space.

\section{Conclusion}
In this work, we make the case for starting with a model capable of filling in the blank when attempting to build a system that can perform both \FitB{} and continuation.
As LMs become bigger, it will be unsustainable to have separately trained models for each generation task.
Multi-task, domain-transferable models, such as the ones we propose, require less total training and are more efficient to store and use at inference time.
While pre-trained T5 by itself is capable of both infilling and continuation, additional conditioning signals such as desired length and goal text can be successfully incorporated into fine-tuning in order to support an even greater diversity of model interactions.
We focused our experiments on the T5 model; however, we expect that other model families and architectures can be trained similarly to support a variety of generation tasks.
For example, GPT-3 \citep{brown2020language} recently began supporting ``insertion'' and ``edit'' interactions in addition to continuation.
Finally, we present a negative result that while few-shot learning is a promising method for
building multi-task support without any finetuning, it is challenging to make work for the \FitB{} task.

\section{Risks and Limitations}
All neural language models, including the ones used in this paper, reflect the biases and other issues present in their training data.
\citet{weidinger2021ethical} discuss these risks in detail.
The models and datasets considered in this paper are all in the English, and the proposed methods may work differently in other languages. 
In addition, the paper mostly focuses on showing results pertinent to the story writing domain; in other domains joint models for continuation and fill-in-the-blank might work worse. 
Finally, the \LLM used in this paper is not publicly available, which to some extent limits reproducibility, though we expect our findings would have been similar had we evaluated with a public model such as GPT-2.
We emphasize that the main contribution of this paper is a comparison of different methods, all of which are easily implementable, rather than new model checkpoints.

\bibliography{acl}

%\pagebreak
\clearpage
\appendix
\renewcommand{\thetable}{A\arabic{table}}
\renewcommand{\thefigure}{A\arabic{figure}}

\section{Appendix}

\subsection{Few-Shot Learning Details}

\label{appendix:fs_prompt_selection}
Choosing appropriate examples for a few-shot prompt can be challenging as task performance is often sensitive to minor changes in prompt design \citep{zhao2021calibrate}.
We experimented with prompts randomly selected from the C4, Reddit Writing Prompts, and ROC Stories training sets, as well as prompts consisting of examples handwritten by the authors with the goal of story-writing in mind.
For each prompt source, we randomly generated five possible prompts, each with three examples.
To simplify the task, we conditioned on desired length but did not include goal conditioning.

\begin{figure}
    \centering
    \includegraphics[width=0.48\textwidth]{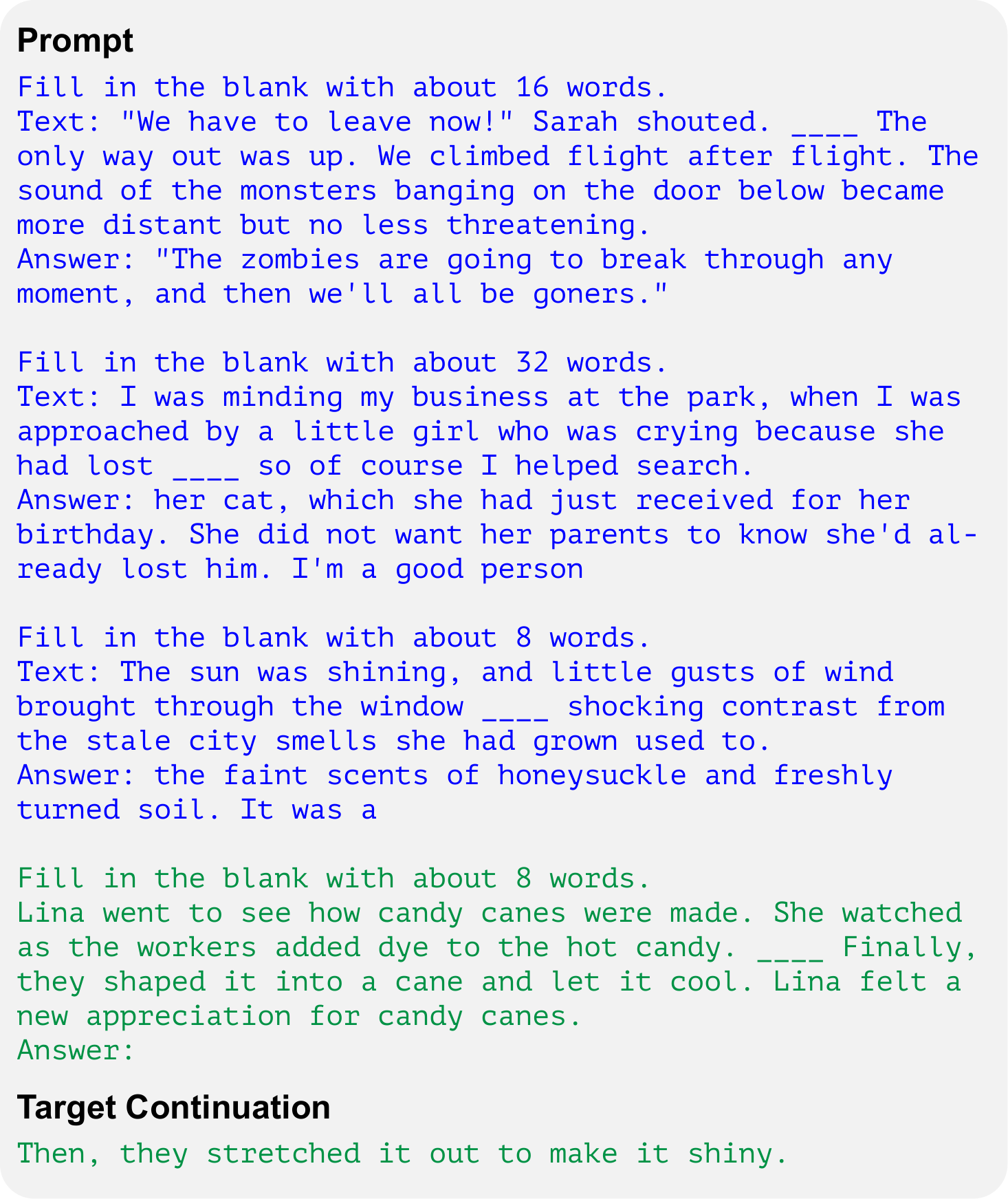}
    \caption{In blue, one of the few-shot prompts that was derived from handwritten examples, and in green, the target example we would like to perform infilling on.}
    \label{fig:few_shot_prompt}
\end{figure}

An example prompt is shown in Figure \ref{fig:few_shot_prompt}.
When choosing random few-shot prompts from the dataset train sets, in order to keep the few-shot prompt text within the 512-token context length limit of the \LLM \citep{thoppilan2022lamda} we used for inference, we only considered examples that contained 100 or fewer tokens, so that the max length of the few-shot prompt was no more than 300 tokens.
This left 212 tokens for the text of the actual example we were interested in performing the \FitB{} task on.
For our hand-written prompt, we wrote the seven examples shown in Table \ref{tab:custom_examples}.
We generated 5 possible prompts by randomly subsampling 3 examples out of these 7 five times.

\begin{figure}
    \centering
    \includegraphics[width=0.49\textwidth]{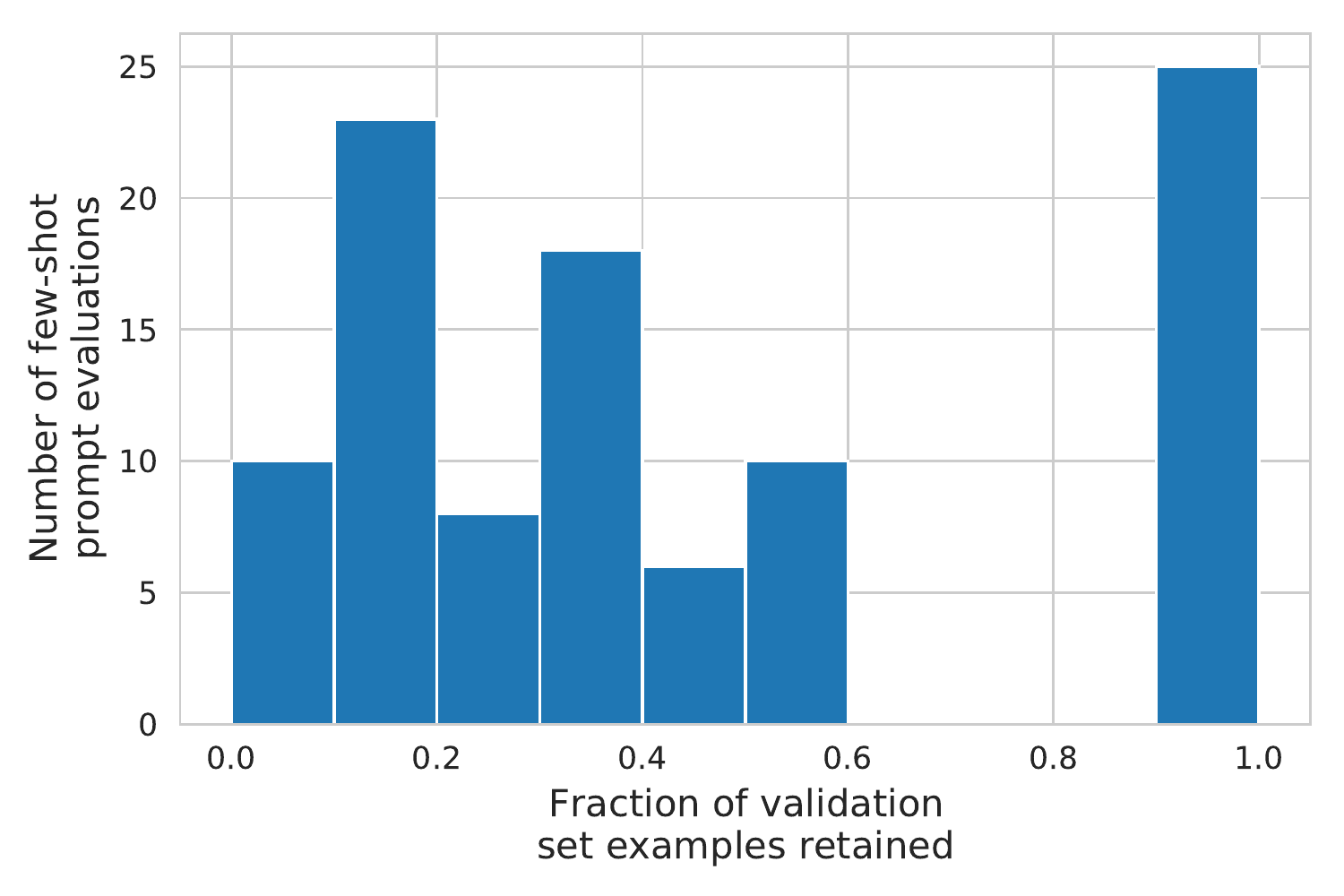}
    \caption{For many of the (validation set, few-shot prompt) combinations, not all validation set examples fit into the maximum sequence length for the \LLM.
    The x-axis on this figure is the fraction of validation set examples which were retained after too-long examples were filtered out.
    The y-axis is the count of (validation set, few-shot prompt) pairs.}
    \label{fig:skipped_fs_examples}
\end{figure}

Table \ref{tab:generative_ppl_results_full} shows the perplexity of the generations from each few-shot prompt.
We note that even leaving room for 212 tokens worth of context text, some evaluation examples did not fit in the prompt length, and these examples were skipped when doing this analysis.
Figure \ref{fig:skipped_fs_examples} shows a histogram of the fraction of validation set examples that remained for each few-shot prompt after the too-long examples were filtered out.
Based on these results, we chose to include in human evaluation the best few-shot prompt from \rocFITB{} and the best few-shot prompt from \cFITB.
Figure \ref{fig:human_eval_results} in the main paper shows the result from the \cFITB{} few-shot prompt, whose outputs were rated slightly higher by human annotators.

Our analysis of few-shot learning prompts was not sufficiently exhaustive to rule out the possibility there might exist a prompt for which this technique would be effective.
For example, we did not conduct formal experiments to systematically vary the prompt wording/formatting shown in Figure \ref{fig:few_shot_prompt}.
What we can conclude is that the process of finding an ideal prompt requires time-consuming trial-and-error and is quite difficult!

\definecolor{lg}{rgb}{0.6,0.6,0.6}

\begin{table}[t]
\small
\centering
    \begin{tabular}{p{7em}|rr}
    % & \multicolumn{2}{c|}{XL} \\
    \toprule
    \textbf{XL Model} & Context & Length \\
     \hline
    \cFITB & 0.867 & 0.810 \\
    \rwpFITB & 0.800 & 0.830 \\
    \arrayrulecolor{lg} \hline \arrayrulecolor{black}
    \cFITE & 0.864 & 0.826 \\
    \rwpFITE & 0.830 & 0.820 \\
    \specialrule{0.8pt}{1pt}{1pt}
    % & \multicolumn{2}{c|}{Large} \\
    \textbf{Large Model} & Context & Length \\
     \hline
    \cFITB & 0.860 & 0.877 \\
    \rwpFITB & 0.797 & 0.881 \\
    \arrayrulecolor{lg} \hline \arrayrulecolor{black}
    \cFITE & 0.858 & 0.775 \\
    \rwpFITE & 0.791 & 0.746 \\
    \bottomrule
    \end{tabular}
    \caption{Accuracy of models finetuned on \FITBFITE{} at correctly using provided length and goal conditioning signals. \label{tab:conditioning_signal_full}} 
\end{table}

\subsection{Experimenting with Prefix Tuning}
During the course of this study, we experimented with the usage of Prefix Tuning \citep{li2021prefix} for the \FitB{} task. In this method, a fixed-length continuous space prefix is appended to the input sequences and this prefix is directly optimized to maximize performance on a given task. This can be used to estimate an upper bound for the performance of few-shot learning on a given task. We trained two prefixes, both of length 5, on pre-trained GPT-2 of size medium (345M) and large (774M) \cite{radford2019language}. While our results showed that the prefix successfully instructed the pre-trained model to perform the \FitB{} task, neither of these models outperformed our few-shot prompts during human evaluation. In fact, they showed only marginally better performance than our random baseline. Due to the discrepancy in parameter count between the prefix tuned GPT-2 models and the \LLM{} model we tested for few-shot prompting, we chose to leave these results out of the final analysis. Future work should seek to explore the limitations of prefix/prompt tuning techniques and the ways in which they and few-shot learning can be fairly compared.

\begin{table*}[htbp]
  \centering
  \small
    \begin{tabular}{l|rrrr}
    \toprule
    %  & \cFITB & \rocFITB & \rwpFITB & \rwpFITB-Sent \\
    & \textsc{C4Fill} & \textsc{RocFill} & \textsc{RwpFill} & \textsc{RwpFill} \\
    \textbf{Few-shot source:} & \textsc{Blank} & \textsc{Middle} & \textsc{Blank} & \textsc{Blank}-Sent \\
    \cline{2-5}
    % \hline
    % \textbf{\cFITB} & 15.67 & \textcolor[rgb]{ 1,  0,  0}{19.72} & \textbf{19.65} & \textbf{16.82} \\
    {\cFITB} & 15.67 & 19.72 & \textbf{19.65} & \textbf{16.82} \\
    {\rocFITB} & \textbf{14.14} & 19.61 & \textbf{19.48} & \textbf{16.36} \\
    {\rwpFITB} & 24.39 & 20.29 & 32.33 & 28.13 \\
    {\rwpFITB-Sent} & 18.91 & \textbf{18.21} & 24.44 & 19.87 \\
    {\textsc{FS Custom}} & 17.98 & 19.80 & 21.72 & 18.38 \\
    \hline
    {Finetuned T5 XL} & 9.99 & 19.00 & 13.64 & 10.03 \\
    Finetuned T5 Large & 10.33 & 20.47 & 14.08 & 10.37 \\
    % \textbf{\FITB{}  Large} & 10.34 & 20.61 & 14.08 & 10.35 \\
    \bottomrule
    \end{tabular}%
\caption{Perplexity of evaluation sets when the blank has been filled in using \LLM{} with few-shot prompting (top) and  our best fine-tuned T5 model (bottom).
% Perplexities are averaged over 5 prompts, and
Among the few-shot results, the best method for each dataset is bolded, as well as methods within one standard error.
}
\label{tab:generative_ppl_results_full}
\end{table*}

\subsection{Finetuning Implementation Details}
\label{appendix:ft_impl_details}
For length conditioning, when discretizing the target sequence's length to a length bucket, we choose the closest value in \{1, 2, 4, 8, 16, 32, 64\} to the target's length in words.

All training was done in the Mesh Tensorflow T5 codebase.\footnote{\url{https://github.com/google-research/text-to-text-transfer-transformer}} Each T5 model was finetuned for about 50,000 steps with a batch size of 128 examples (i.e., $\sim$6.4M examples were seen during finetuning.)
A constant learning rate of 0.0008 was used, and no overfitting was observed. 

\subsection{Further Finetuning Experiments}
\label{appendix:further_ft_exp}
In the main paper, we focused on a single finetuning setting, one where half the examples have randomly placed blanks and the other half have blanks always at the end.
We actually experimented with three possible finetuning settings:
\begin{itemize}[noitemsep,nolistsep]
    \item In the standard \FITB{} setting, the blank location is sampled uniform randomly across the sequence.
    \item In the \FITBFITE{} setting, for half of the examples the blank is randomly selected and for the other half it is always at the end. As we hypothesized that finetuning on such data would result in better performance at the continuation task, this was the setting we used in the main paper.
    \item In the \FITE{} (a.k.a. \textsc{Lm-Adaption}) setting, the blank is always placed at the end of the sequence. In essence, we are finetuning solely for the continuation objective. 
\end{itemize}

For the \FITBFITE{} setting from the main paper, we additionally experimented with finetuning a 3B parameter ``XL'' T5 model.

Table \ref{tab:ft_perplexity_results} shows the perplexity of all these models on a variety of validation sets.
Note that these are perplexities in the conventional definition--perplexity of the target sequence given the input sequence using examples from the validation set--not the fluency measure we report in the main paper.

The perplexity numbers across the different models are comparable, since all models used the default T5 vocabulary.
The perplexity numbers across different datasets are not comparable since some datasets, like ROC Stories, are simply easier to model than others.
Unsurprisingly, the larger models achieved lower perplexity on all validation sets.
We can also see from Table \ref{tab:ft_perplexity_results} that it was probably not strictly necessary to enforce that 50\% of training examples had blanks at the end.
The model finetuned exclusively with randomly placed blanks (\FITB{}) performed only slightly worse (probably not statistically significant) on the continuation-style validation sets than the \FITBFITE-trained model.

\definecolor{lg}{rgb}{0.6,0.6,0.6}

\begin{table*}
\small
\centering
    \begin{tabular}{l|l|cc|cc|ccc}
    \toprule
    \multicolumn{1}{c}{Pre-trained} & & \multicolumn{2}{c}{\textsc{C4Fill}} & \multicolumn{2}{c}{\textsc{RocFill}} & \multicolumn{3}{c}{\textsc{RwpFill}}\\ 
    \multicolumn{1}{c}{model} & \multicolumn{1}{c|}{Finetune setting} & \multicolumn{1}{c}{\textsc{Blank}} & \multicolumn{1}{c}{\textsc{End}} & \multicolumn{1}{c}{\textsc{Middle}} & \multicolumn{1}{c}{\textsc{End (T)}} &
    \multicolumn{1}{c}{\textsc{Blank}} & \multicolumn{1}{c}{\textsc{SentBlank}} &
    \multicolumn{1}{c}{\textsc{End}}\\
    \hline
    T5 Large & \FITBFITE & 11.79 & 13.47 & 6.43 & \textbf{6.73} & 16.15 & \textbf{14.84} & \textbf{19.89} \\
    T5 Large & \FITB & \textbf{11.64} & 13.88 & \textbf{6.41} & 6.84 & \textbf{16.11} & 14.89 & 20.16 \\
    T5 Large & \FITE & 16.10 & \textbf{13.26} & 37.08 & 6.79 & 21.35 & 27.73 & 19.90 \\
    \hline
    T5 XL & \FITBFITE & {9.53} & {11.15} & {5.34} & {5.79} & {13.05} & {11.98} & {16.57} \\
    \bottomrule
    \end{tabular}
    \caption{The perplexity according to T5 Large finetuned with three possible training data settings, with blanks placed randomly (\FITB), with blanks placed always at the end (\FITE), or with  an equal mix of these two (\FITBFITE).
    For the large-sized models, the one that achieved lowest perplexity on each dataset is bolded.
% Except for \textsc{RocFillEnd-S5-F}, where we would like the model to assign high perplexity to stories with incorrect endings, lower is better.
\label{tab:ft_perplexity_results}} 
\end{table*}

Finally, Table \ref{tab:conditioning_signal_full} shows the accuracy of both model sizes on the two conditioning signals which were incorporated: length bucket and goal conditioning.
Surprisingly, using a larger model improves goal conditioning accuracy but hurts length conditioning accuracy.

\subsection{Further Human Evaluation Details}
\begin{figure}
    \centering
    \includegraphics[width=0.49\textwidth, frame]{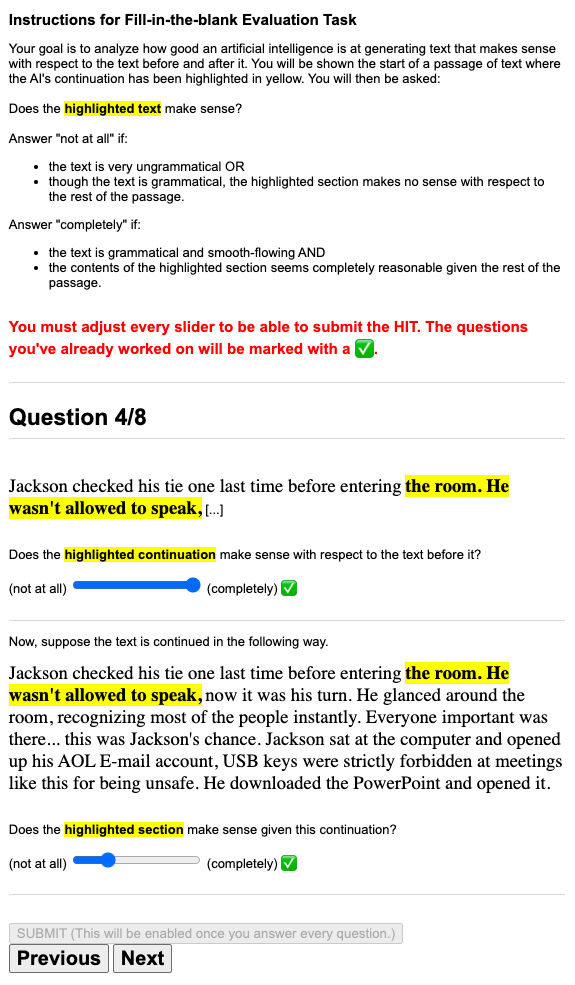}
    \caption{A screenshot of the question structure for human evaluation.}
    \label{fig:amturk_ui}
\end{figure}

A screenshot of the Human Intelligence Task (HIT) used for annotations is shown in Figure \ref{fig:amturk_ui}. Workers were paid originally paid \$1.85 per HIT, but since the average HIT duration ended up being 15 minutes, we awarded each rater a bonus to raise their pay to an average of \$10 per hour.
We restricted the HITs to workers for whom Masters had been granted and who had previously done at least 100 HITs.

Each example was shown to three raters, and annotations were rejected if the rater gave a lower overall score to the random output than to the ground-truth one.
A total of 3 annotations were rejected.
Overall, the Fleiss' kappa agreement of pairs of annotators giving the same numerical score to the same question was 0.26.

\begin{table*}[]
    \centering
    \small
    \begin{tabular}{|p{0.5\linewidth}|p{0.3\linewidth}|}
         \hline
         \textbf{Context} & \textbf{Target} \\
         \hline
An elderly man was sitting alone on a dark path. The man looked down at his feet, and realized \_\_\_\_ . It was a plain pine box and looked as if it had been there for a long time. The man was afraid to look inside the box. & he was holding a bright red box made of pine \\
\arrayrulecolor{lg} \hline \arrayrulecolor{black}
The mantle was cluttered with objects: \_\_\_\_ and more than one vase of dried flowers. The bejeweled lamp was at the very back, nearly invisible. & picture frames showing grandchildren and long-ago weddings, knickknacks collected from all over the world, \\
\arrayrulecolor{lg} \hline \arrayrulecolor{black}
"We have to leave now!" Sarah shouted. \_\_\_\_ The only way out was up. We climbed flight after flight. The sound of the monsters banging on the door below became more distant but no less threatening. & "The zombies are going to break through any moment, and then we'll all be goners." \\
\arrayrulecolor{lg} \hline \arrayrulecolor{black}
The sun was shining, and little gusts of wind brought through the window \_\_\_\_ shocking contrast from the stale city smells she had grown used to. & the faint scents of honeysuckle and freshly turned soil. It was a \\
\arrayrulecolor{lg} \hline \arrayrulecolor{black}
I was minding my business at the park, when I was approached by a little girl who was crying because she had lost \_\_\_\_ so of course I helped search. & her cat, which she had just received for her birthday. She did not want her parents to know she'd already lost him. I'm a good person \\
\arrayrulecolor{lg} \hline \arrayrulecolor{black}
It was a cold night, and a storm was raging out at sea. A lightning bolt lit up the sky, briefly illuminating the lighthouse \_\_\_\_ plummeted but just before reaching the churning water, he disappeared in a poof of purple flame! & and the young man peering hesitantly over the sheer cliff. Before the next peal of thunder he jumped. At first he  \\
\arrayrulecolor{lg} \hline \arrayrulecolor{black}
The magician pulled out of his pocket \_\_\_\_ and then a second one and a third. He didn't stop until soon the ground was covered with them. & a scarlet handkerchief  \\
         \hline
    \end{tabular}
    \caption{Hand-written fill-in-the-blank examples used for ``custom'' prompt during few-shot learning.}
    \label{tab:custom_examples}
\end{table*}

\end{document}